\documentclass{article} 
\usepackage{amsmath}
\usepackage{nips13submit_e,times}
\usepackage{hyperref}
\usepackage{url}

\title{Continuous Learning: Engineering Super Features With Feature Algebras}

\author{
Michael Tetelman\\
Advanced Magic Technologies\\
Los Gatos, CA 95032, USA\\
\texttt{michael.tetelman@gmail.com} \\
}

\newenvironment{itemizeminus}{\begin{itemize} }{\end{itemize}}

\DeclareMathOperator*{\argmax}{arg\,max}

\nipsfinalcopy 

\begin{document}

\maketitle

\begin{abstract}
In this paper we consider a problem of searching a space of predictive models 
for a given training data set.
We propose an iterative procedure for deriving a sequence of improving models 
and a corresponding sequence of sets of non-linear features on the original input space.
After a finite number of iterations $N$, the non-linear features become $2^N$-degree 
polynomials on the original space.
We show that in a limit of an infinite number of iterations derived non-linear features 
must form an associative algebra: a product of two features is equal to a linear 
combination of features from the same feature space for any given input point.
Because each iteration consists of solving a series of convex problems that contain all previous solutions,
the likelihood of the models in the sequence is increasing with each iteration  
while the dimension of the model parameter space is set to a limited controlled value.
\end{abstract}

\section{Introduction}

This paper proposes a method of finding a sequence of improving models
for a given fixed data set that we call Continuous Learning. 
The method is based on iterative exploration of a space of models that have a specific limited number of parameters, 
which correspond to non-linear polynomial features of an input space.
The feature set is evolving with each iteration, so the most important features are selected from the current feature set 
then the reduced feature set is expanded to include higher degree polynomials 
while the dimension of the expanded feature space is limited or fixed.
Resulting features are computed recursively from iteration to iteration with different parameters of the recursions found 
for each execution of the iteration algorithm.

The paper is organized as follows: 

To search a model space we need to compare different models and solutions.
We use Bayesian framework that provides a natural way for model comparison \cite{barberBRML2012}.
Continuous Learning consists of a sequence of iteration cycles.
Each iteration cycle has a number of steps. In first of these steps, we use bootstrap method 
to obtain a set of sampled solutions \cite{hastie_09_elements-of.statistical-learning} and parameter-feature duality 
as a method for exploring the feature space. We construct a model of a solution 
distribution and use Principal Component Analysis to select a subspace of the most important 
solutions and reduce dimensions of the parameter space. Then we use non-linear expansion 
of the feature space by adding tensor-features that are products of the principal features 
selected in the previous step. That concludes a definition of one iteration cycle.

Each iteration is recursively redefining features that become non-linear functions on the original feature space.
We analyzed a stationary solution of the iteration cycle and found that, in a limit of the infinite number of iterations, 
features form a Feature Algebra. Different solutions of Feature Algebra define non-linear feature 
representations.

For the purpose of this work, we will consider a prediction problem setup.
The goal is to find a probability of a class label $y$ for a given input $x$:
$Prob (y|x)$.

The model class family is defined by a conditional probability function $P
(y|x, w)$, where $w$ is a parameter vector. For a set of independent training
data samples $(y_t, x_t), t = 1 \ldots t_{\max}$, the probability of labels for
given inputs is defined by the Bayesian integrals

\begin{equation} \label{eq:BayesianIntegral1}
    B (y, \{y_t \}|x, \{x_t \}) = \int P (y|x, w) \prod_t P (y_t |x_t, w)  P_0 (w|r) dw, 
\end{equation}
\begin{equation} \label{eq:BayesianIntegral2}
    B (\{y_t \}|\{x_t \}) = \int \prod_t P (y_t |x_t, w)  P_0 (w|r) dw, 
\end{equation}
\begin{equation} \label{eq:PredictionProb}
    Prob (y|x) = B (y, \{y_t \}|x, \{x_t \}) / B (\{y_t \}|\{x_t \}), 
\end{equation}

where $P_0 (w|r)$ is a prior probability of parameters $w$ that guarantees existence
and convergence of the Bayesian integrals in Equations \ref{eq:BayesianIntegral1} and \ref{eq:BayesianIntegral2} 
and it is normalized as follows:

\begin{equation} \label{eq:PriorProbNormalization}
    \int P_0 (w|r) dw = 1. 
\end{equation}

Then the Bayesian integral in the Equation \ref{eq:BayesianIntegral2} is equal to a probability of labels $\{y_t\}$ 
for the given input vectors $\{x_t\}$ of the training set.

The prior distribution $P_0(w|r)$ itself depends on some parameters $r$ (hyper-parameters or regularization parameters). 
Then the Bayesian integrals in the Equations \ref{eq:BayesianIntegral1} and \ref{eq:BayesianIntegral2} also depend 
on regularization parameters $r$. Optimal values of the regularization 
parameters could be found by maximizing the probability of the training data given by the 
Bayesian integral in Equation \ref{eq:BayesianIntegral2}. This is possible because due to normalization of the prior probability in the Equation
\ref{eq:PriorProbNormalization}, the Bayesian integrals include the contribution of the normalization factor that 
depends only on regularization parameters $r$. The regularization solution found by maximizing the probability 
of training data is equivalent to the solution for regularization parameters found by cross-validation. 

For the purpose of this paper, it is sufficient to estimate values of the Bayesian integrals in the Equations 
\ref{eq:BayesianIntegral1} and \ref{eq:BayesianIntegral2} using maximum likelihood approximation (MLA) 
by finding a solution $w_m$ that maximizes log-likelihood of the training set that includes the prior probability of parameters $w$:

\begin{equation} \label{eq:MLAsolution}
    w_m = \argmax_w L(w), L (w) = \left\{ \ln (P_0 (w)) + \sum_t \ln (P(y_t |x_t, w)) \right\} .
\end{equation}

Then the Bayesian integral can be approximated by the maximum likelihood as follows

\begin{equation} \label{eq:MLAintegral}
    B (\{y_t \}|\{x_t \}) \approx \exp (L (w_m)). 
\end{equation}

\section{Training and sampling noise}

The training data are always a limited-size set or a selection from a
limited-size set. For that reason it contains a sampling noise. The sampling
noise affects solutions. This is easy to see by sub-sampling the training data: for
each sampled training set the MLA solution is different in the Equation \ref{eq:MLAsolution} 
as well as the value of the Bayesian integral $B$ in the Equation \ref{eq:MLAintegral}.

Our goal is to find a solution that is the least dependent on the sampling noise
and better represents an actual statistics of a source.

To achieve that, we can sample the training data to create a set of the training
sets $\left\{ T_s, s = 1 \ldots 2^{t_{\max}} \right\}$ and then find an MLA
solution $w_s$ for each sampled training set $\left\{ y_t, x_t, t \in T_s
\right\}$. 

\begin{equation} \label{eq:MLAsolutionS}
    w_s = \argmax_w \left\{ \ln (P_0 (w)) + \sum_{t \in T_s} \ln (P  (y_t |x_t, w)) \right\},   B_s \approx \exp ( L(w_s) ) . 
 \end{equation}

Now we have a set of solutions $\left\{ w_s \right\} $ which is a noisy representation of a source statistics. 
The probability of each solution is given by the value of the Bayesian integral on the solution and is equal to

\begin{equation} \label{eq:SolutionProb}
    Prob(w_s) = e^{L \left( w_s \right)}  / \sum_{s'} e^{L \left( w_{s'} \right)} .
\end{equation}

The solution distribution in the Equation \ref{eq:SolutionProb} is based on sampling 
of the original training data set. It is a variant of a bootstrap technique 
\cite{hastie_09_elements-of.statistical-learning}. This type of methods 
is actively used in different forms to improve training and to find a robust solution, 
that is less dependent on a sampling noise in the original training set. For example, see  
the dropout method that randomly omits hidden features during training \cite{DBLP:journals/corr/abs-1207-0580} 
or the method of adding artificial noise or a corruption to a training data \cite{DBLP:journals/corr/abs-1305-6663}.

Instead of trying to find a single best solution we use the bootstrap method here to obtain a distribution of solutions.

\section{Distribution of solutions}

Let's consider the set of solutions in the Equation \ref{eq:MLAsolutionS} as samples from unknown distribution
of solutions, where each solution $w_s$ has a weight $\exp \left( L \left( w_s
\right) \right)$ and the probability of the solution is given by the Equation \ref{eq:SolutionProb}. 

Then we can model this distribution of solutions by
proposing a probability function $Prob \left( w_s \left| z \right.
\right)$ with parameters $z$, which we can find by maximizing by $z$ the following log-likelihood

\[
    \sum_s e^{L \left( w_s \right)} \ln \left( Prob \left( w_s \left| z \right. \right) \right) / \sum_s e^{L \left( w_s \right)} .
\]

Up until now we did not specify the model class distribution $P
\left( y \left| x, w \right. \right)$. For the following consideration, we will use logistic
regression for model class distribution with binary label $y = 0, 1$ as follows

\begin{equation} \label{eq:modelLogit}
    P \left( y \left| x, w \right. \right) = \frac{\exp \left( y\mathbf{w}
   \cdot \mathbf{f} \left( x \right) \right)}{1 + \exp \left( \mathbf{w}
   \cdot \mathbf{f} \left( x \right) \right)} , 
\end{equation}
   
where $\mathbf{w} \cdot \mathbf{f} \left( x \right) = w_0 + \sum_i w_1^i
x^i$ is a product of the parameter vector $\mathbf{w}$ with the feature vector
$\mathbf{f} \left( x \right)$ which includes a bias feature 1.

The use of the logistic regression here is not a limitation on the possible models. 
It is selected only for certainty and to avoid an unnecessary complication of the consideration. 
As we will see the following approach is applicable to any model class distribution that is a function of a scalar product of a feature vector
and a parameter vector. Also it could be used for a model class distribution that is a function of multiple products 
of a feature vector and parameter vectors.

Using the Equation \ref{eq:modelLogit} we will find a set of solutions defined in the Equation \ref{eq:MLAsolutionS} 
for each corresponding training set $ T_s $.

To model the distribution of solutions we will start by considering Gaussian model 
for the distribution of solutions $Prob\left( w_s \left| z \right. \right)$. 

Then the model is defined by mean

\begin{equation} \label{eq:meanSolution}
    \langle w \rangle = \sum_s w_s e^{L \left( w_s \right)} / \sum_s e^{L \left( w_s  \right)}
\end{equation}

and covariance matrix

   \[ cov \left( w \right) = \langle \left( w - \langle w \rangle \right) \otimes \left( w - \langle w \rangle \right) \rangle = \]
\begin{equation} \label{eq:covSolution}
    \sum_s \left( w_s - \langle w \rangle \right) \otimes \left( w_s - \langle w \rangle \right) e^{L \left( w_s \right)} / \sum_s e^{L \left( w_s \right)} . 
\end{equation}

We will use Principal Component Analysis (PCA) to separate important solutions from noise. 
That leads to the following representation of the parameter vector $w$:

\begin{equation} \label{eq:PCAparameters}
    \mathbf{w} = \mathbf{V_0} + \sum_{\alpha} V_1^{\alpha} \mathbf{U_{\alpha}}, 
\end{equation}

where $\mathbf{V_0} = \langle \mathbf{w} \rangle$ and $\mathbf{U_{\alpha}}$ are selected eigenvectors of the covariance
matrix $cov \left( w \right)$ indexed by $\alpha$. The coordinates
$V_1^{\alpha}$ span over the principal-component subspace that is defined by selected
eigenvectors. The selected eigenvectors correspond to a high-variance subspace,
where eigenvalues of the covariance matrix $cov \left( w \right)$ are
larger than a certain threshold. The value of the threshold for selecting
the principal components is a hyper-parameter that controls the dimension of the principal
component space, which in practice is constrained by the available memory.

\section{Iterating over sequence of models}

The important property of the model class probability distribution is a 
parameter - feature duality: 
the parameters $w$ for the model class distribution $P \left( y \left| x, w \right. \right)$ 
are used only in a product form

\begin{equation} \label{eq:origProduct}
    \mathbf{w} \cdot \mathbf{f} \left( x \right) = w_0 + \sum_i w_1^i x^i, 
\end{equation}

where $\mathbf{f} \left( x \right)$ are the original features.

By considering solutions that are limited to the principal component space we can
find that the product is given by the following Equation

\begin{equation} \label{eq:parFeatureProd}
    \mathbf{w} \cdot \mathbf{f} \left( x \right) = \left( \mathbf{V}_0
   \cdot \mathbf{f} \left( x \right) \right) + \sum_{\alpha} V_1^{\alpha}
   \left( \mathbf{U}_{\alpha} \cdot \mathbf{f} \left( x \right) \right) .
\end{equation}

We will have exactly the same product form here as in the Equation \ref{eq:origProduct} when we will define new features
$F_0(x), F_{\alpha}(x)$ via original features $\mathbf{f}(x)$ as follows

\begin{equation} \label{eq:reDefinedFeatures}
    F_0 \left( x \right) = \left( \mathbf{V}_0 \cdot \mathbf{f} \left( x
   \right) \right)/| \mathbf{V}_0 | , F_{\alpha} \left( x \right) =
   \left( \mathbf{U}_{\alpha} \cdot \mathbf{f} \left( x \right) \right),
\end{equation}

so the parameter-feature product will look like this

\begin{equation} \label{eq:newProduct}
    \mathbf{w} \cdot \mathbf{f} \left( x \right) = V_0 F_0 \left( x \right) +
   \sum_{\alpha} V_1^{\alpha} F_{\alpha} \left( x \right), 
\end{equation}

where now $V_0$ and $V_1^{\alpha}$ are new parameters for the model class distribution with
re-defined super-features $F_0(x), F_{\alpha} \left( x \right)$ from the Equation \ref{eq:reDefinedFeatures}. 

The result of this step is that using PCA and redefining features we reduced 
the original parameter space to a new smaller space.

Let's now extend the parameter-feature space by adding products of
the super-features

\begin{equation} \label{eq:extendingFeatures}
    \mathbf{w} \cdot \mathbf{f}(x) \rightarrow V_0 F_0(x) 
    + \sum_{\alpha} V_1^{\alpha} F_{\alpha}(x) + V_2^{0 0}F_0(x)^2 
    + \sum_{\alpha} V_2^{0 \alpha} F_0(x) F_{\alpha}(x) 
    + \sum_{\alpha, \beta} V_2^{\alpha \beta} F_{\alpha}(x) F_{\beta}(x). 
\end{equation}

By extending the feature space in the Equation \ref{eq:extendingFeatures}, we increased the dimension 
of the parameter space by adding new parameters $ V_2 $ and 
creating new features as non-linear (quadratic) functions of the previous features.

Now we can repeat the iteration cycle, which consists of the steps in the Table \ref{tab:iteration}.

\begin{table}[h]
\centering
\begin{tabular}{| l l |} 
    \hline
    & \\
  - & sample data   \\
  - & get solution set  \\
  - & select principal components   \\
  - & redefine features \\
  - & extend feature set by adding products of features. \\
    & \\
    \hline
\end{tabular}
\caption{Iteration cycle}
\label{tab:iteration}
\end{table}

It is important to emphasize that 

\begin{itemizeminus}
\item At each iteration we have a model that is defined on a new feature space and has 
a limited defined number of dimensions in its parameter space. 

\item At each iteration the feature space is a non-linear map of the original feature space.

\item Each iteration makes new super-features to be higher degree polynomials of the original basic features. 

\item After $N$ iterations new features are $2^N$-degree polynomials of the original features.

\end{itemizeminus}

The expansions of the feature set by adding products of features were used in recently proposed 
sum-product networks \cite{DBLP:journals/corr/abs-1202-3732} and 
Neural Tensor Networks \cite{DBLP:journals/corr/abs-1301-3618}.

\section{Feature Algebra}

To simplify notations, let's allow the feature indices $\alpha, \beta$ to include value $0$. Then the Equation \ref{eq:extendingFeatures} 
will look like this

\begin{equation} \label{eq:extendingFeaturesNewAlpha}
    \mathbf{w} \cdot \mathbf{f}(x) \rightarrow \sum_{\alpha} V_1^{\alpha} F_{\alpha}(x) 
    + \sum_{\alpha, \beta} V_2^{\alpha \beta} F_{\alpha}(x) F_{\beta}(x). 
\end{equation}

The iterations will converge when the product of super-features $F_{\alpha} \left(
x \right)$ in the Equation \ref{eq:extendingFeaturesNewAlpha} could be expressed only as a linear combination of the super-features

\begin{equation} \label{eq:fAlgebra}
    F_{\alpha} \left( x \right) F_{\beta} \left( x \right) = \sum_{\gamma}
   C_{\alpha \beta}^{\gamma} F_{\gamma} \left( x \right), 
\end{equation}

where for $\alpha=0$ the super-feature $F_{\alpha}(x)$ is the bias super-feature $F_0(x)$.

The Equation \ref{eq:fAlgebra} defines a feature algebra with structure constants $C_{\alpha
\beta}^{\gamma}$.

The feature algebra has following important properties:

\begin{enumerate}

  \item It must be associative: 
  
  \begin{equation}
  F_{\alpha} \left( x \right) \left( F_{\beta}
  \left( x \right) F_{\gamma} \left( x \right) \right) = \left( F_{\alpha}
  \left( x \right) F_{\beta} \left( x \right) \right) F_{\gamma} \left( x
  \right), 
  \end{equation}
  
  that property leads to major equations for structure constants:

  \begin{equation} \label{eq:structureConstants}
    \sum_{\mu} C_{\alpha \beta}^{\mu} C_{\mu \gamma}^{\nu} = \sum_{\mu}
     C_{\alpha \mu}^{\nu} C_{\beta \gamma}^{\mu};
  \end{equation}

  \item The super-feature space with the feature algebra is a complete linear vector
  space: due to the algebra, any function $g(F)$ on the super-feature space representable
  by power series is equal to a linear combination of the super-features with computable coefficients $A_{\alpha}$:

  \begin{equation} \label{eq:falgSpaceLinearity}
    g \left( F \left( x \right) \right) = \sum_{\alpha} A_{\alpha}
     F_{\alpha} \left( x \right) . 
  \end{equation}
  
\end{enumerate}

The feature algebra defined by the Equation \ref{eq:fAlgebra} is not limited to 
polynomial functions, it could be any function set that satisfies the algebra Equation \ref{eq:fAlgebra} with structure constants that 
are a solution of the Equation \ref{eq:structureConstants}.

Simple examples of algebras that are defined by Equations \ref{eq:fAlgebra} and 
\ref{eq:structureConstants} are complex numbers and quaternions.
Less trivial examples of such algebras are operator algebras that were 
successfully used in Statistical Physics of Phase Transitions and Quantum Field Theory.

\section{Conclusions}

We proposed an iterative procedure for generating non-linear features
(super-features) that are high-degree polynomials on the original feature space
after a finite number of iterations.

For a finite number of iterations, the non-linear super-features are defined by
sets of principal components selected at each iteration.

By selecting a small set of principal components, the dimensionality of a feature space is limited
at each iteration while resulting super-features are highly non-linear
(as polynomials of exponentially high with number of iterations degree). That contrasts with an approach
when high-degree polynomials are used as the original features - which requires
to find a solution for an exponentially high-dimensional model.

In the limit of infinite iterations, the super-features form a linear vector space
with an associative algebra.

\subsubsection*{Acknowledgments}

I am grateful to my wife Yelena for support.

\small{

\providecommand{\bysame}{\leavevmode\hbox to3em{\hrulefill}\thinspace}
\providecommand{\MR}{\relax\ifhmode\unskip\space\fi MR }
\providecommand{\MRhref}[2]{%
  \href{http://www.ams.org/mathscinet-getitem?mr=#1}{#2}
}
\providecommand{\href}[2]{#2}


}

\end{document}